\documentclass{article}
\usepackage{spconf,amsmath,graphicx}
\usepackage[hyphens]{url}
\usepackage{amsmath}
\usepackage{paralist} 
\usepackage{booktabs}
\usepackage[dvipsnames]{xcolor}
\usepackage{microtype}
\usepackage{ragged2e}
\newcommand{\rulesep}{\unskip\ \vrule\ }
\PassOptionsToPackage{hyphens}{url}
\usepackage{caption}
\usepackage{subcaption}
\RequirePackage{xcolor}
\usepackage{multirow}
\usepackage[hyphens]{url}
\usepackage[pagebackref=false,breaklinks=true,colorlinks,bookmarks=false]{hyperref}

\usepackage{color}

\usepackage{enumitem}

\begin{document}

\title{Eyes Tell All: Irregular Pupil Shapes Reveal GAN-generated Faces}


\name{Hui Guo$^{1}$, Shu Hu$^{2}$, Xin Wang$^{2,3}$, Ming-Ching Chang$^{1}$, Siwei Lyu$^{2}$}
\address{
$^{1}$University at Albany, State University of New York, USA. ~~~  
 {\tt\{hguo,mchang2\}@albany.edu} \\
$^{2}$University at Buffalo, State University of New York, USA. ~~~  
 {\tt\{shuhu,siweilyu\}@buffalo.edu}   \\
$^{3}$Keya Medical, Seattle, Washington, USA.  ~~~   
{\tt xwang264@buffalo.edu,xinw@keyamedna.com} 
}

\maketitle

\begin{abstract}
Generative adversarial network (GAN) generated high-realistic human faces are visually challenging to discern from real ones. They have been used as profile images for fake social media accounts, which leads to high negative social impacts. 
In this work, we show that GAN-generated faces can be exposed via irregular pupil shapes. This phenomenon is caused by the lack of physiological constraints in the GAN models. 
We demonstrate that such artifacts exist widely in high-quality GAN-generated faces. We design an automatic method to segment the pupils from the eyes and analyze their shapes to distinguish GAN-generated faces from real ones. 
Qualitative and quantitative evaluations of our method on the Flickr-Faces-HQ dataset and a StyleGAN2 generated face dataset demonstrate the effectiveness and simplicity of our method.
\end{abstract}

\begin{keywords}
image forensics, GAN faces detection, pupil segmentation, fake face detection
\end{keywords}


\section{Introduction}
\label{sec:intro}
\vspace{-0.2cm}


The rapid development of the generative adversarial network~(GAN) models~\cite{karras2017progressive,karras2019style,karras2020analyzing} has made it possible to synthesize highly realistic human face images that are difficult to discern from real ones~\cite{nightingale2021synthetic}. These GAN-generated faces have been misused for malicious purposes. Recent years have seen an increasing number of reports that GAN-generated faces were used as profile images on fake social media accounts, which generates negative social impacts~\cite{theverge,cnn1,cnn2,reuters}.

\begin{figure}[t]
\centering

\begin{subfigure}[b]{0.45\textwidth}
  \includegraphics[width=\linewidth]{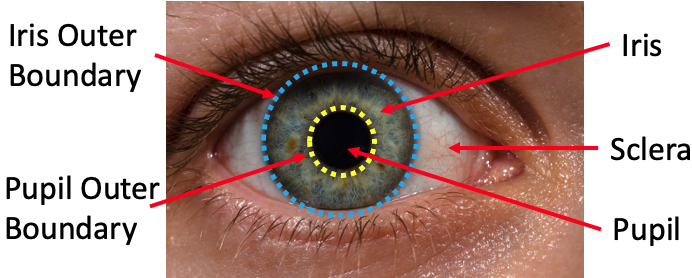}
  \label{fig:mnist_noise_20}
\end{subfigure}
\vspace{-1.5em}
\begin{subfigure}[b]{0.48\textwidth}
  \includegraphics[width=\linewidth]{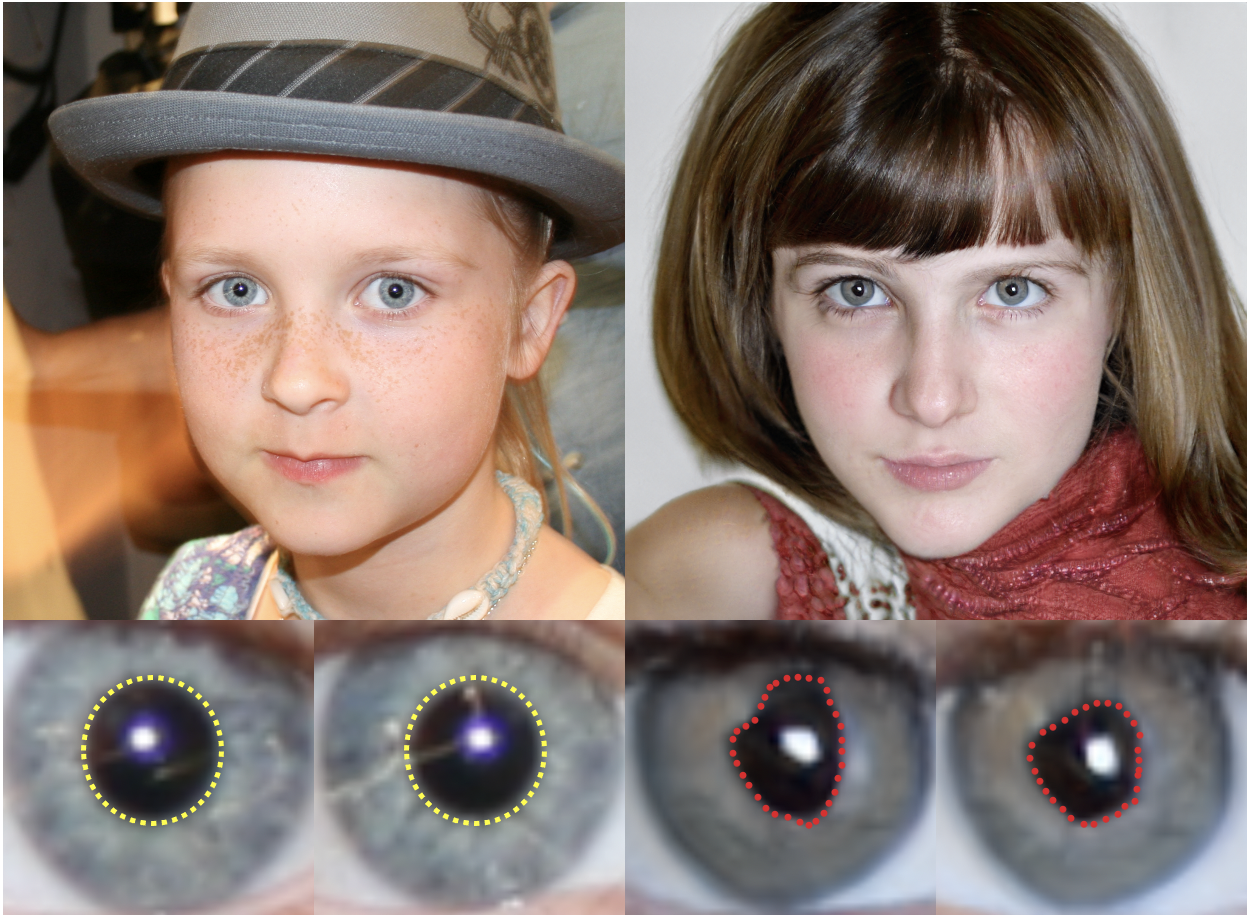}
  \label{fig:mnist_noise_30}
\end{subfigure}%
\vspace{-2em}
\caption{\em 
\textbf{Top:} Anatomy of a human eye: the iris and pupil are at the center surrounded by the sclera. \textbf{Bottom:} Examples of pupils of real human (left) and GAN-generated (right). Note that the pupils for the real eyes are in circular or elliptical shapes (\textbf{yellow}), while those for the GAN-generated pupils are much irregular (\textbf{red}). For GAN-generated faces, the shapes of the pupils are very different from each other when zoomed-in. 
}
\vspace{-1em}
\label{fig:compare}
\vspace{-1em}
\end{figure}


Such pernicious impact of these fake faces has led to the development of methods aiming to distinguish GAN-generated images from real ones. 
Many of those methods are based on deep neural network (DNN) models due to their high detection accuracy~\cite{marra2019gans,wang2020cnn,guo2021robust}. Albeit such success, these methods suffer from two significant limitations: (1) the lack of interpretability of the detection results and (2) the low capability to generalize across different synthesis methods \cite{hu2020learning, hu2021sum}. 

Another category of GAN-generated image detection methods aims to expose the inadequacy of the GAN models in handling the physical constraints of the face representation and synthesis process~\cite{yang2019exposinggan,yang2018exposing,li2018detection,matern2019exploiting}. Since these methods exploit knowledge of the physical world in distinguishing the fake images from real ones, these methods are more interpretable and can work robustly against various synthesis methods. The recent work \cite{hu2021exposing} exploits the inconsistency of the corneal specular highlights between the two synthesized eyes to identify GAN-generated faces. 
However, this method is limited by its environmental assumptions regarding the light sources and the reflectors that must be visible in both eyes. This might cause high false negatives in fake face detection.  






In this work, we explore a universal physiological cue of the eye, namely the {\em pupil shape consistency}, to reliably identify GAN-synthesized faces. 
As shown in Figure~\ref{fig:compare}, the eye is one of the few organs in the human body that is highly circular and regular in geometry. We hypothesize that the human iris and pupil can provide rich physical and physiological cues that can improve GAN-synthesized face detection. 

Our method is based on a simple physiological assumption that human pupils should be nearly circular in their shapes in a face image. Due to different facial orientations and camera angles, the actual pupil shapes can be elliptical. Our observation is that {\em this simple property is not well preserved in the existing GAN models}, including StyleGAN2~\cite{karras2020analyzing}, the state-of-the-art face synthesis method. 
The pupils for the StyleGAN-generated faces tend to have non-elliptical shapes with irregular boundaries. Figure~\ref{fig:compare} shows an example with zoom-in views of the pupils. 
Such artifacts in the GAN-generated faces are due to the difficulty or negligence of physiological constraints on human anatomy when training the GAN models via standard data-driven machine learning.

The proposed GAN-generated face detector consists of several automatic steps. We first segment the pupil regions of the eyes and extract their boundaries automatically. 
We next fit an ellipse parametric model to each pupil, and calculate the Boundary Intersection-over-Union (BIoU) scores~\cite{cheng2021boundary} between the predicted pupil mask and the ellipse-fitted model. The BIoU score provides a quantitative measure of the regularity of the pupil shape, that determines if the eyes (and the face) are real or not. Experiments are conducted on a dataset containing both real and machine-synthesized faces. Results in $\S$~\ref{sec:experiments} show that there is a clear separation between the distributions of the BIoU scores of the real and GAN-generated faces. 

The main contributions of this work are two-fold:
\vspace{-0.2cm}
\begin{itemize}[leftmargin=16pt] \itemsep -.2em

\item We are the first to propose the idea of exploiting pupil shape consistency as an effective way to distinguish fake faces from real ones. This new cue is effective for humans as well to visually identify GAN-generated faces \cite{guo2022open}.

\item The proposed method for fake face detection is based on explainable physiological cues. It is simple, effective, and explainable. Evaluations on the Flickr-Faces-HQ dataset and an in-house collected StyleGAN2 face dataset show its effectiveness and computational efficiency. 

\end{itemize}





\section{Related Works}
\label{sec:related}
\vspace{-0.2cm}

\textbf{GAN-generated faces.} A series of recent GAN models have demonstrated superior capacity in generating or synthesizing realistic human faces. 
However, the works~\cite{yang2019exposinggan, matern2019exploiting} indicate that faces generated by the early StyleGAN model~\cite{karras2019style} have considerable artifacts such as fingerprints~\cite{marra2019gans, yu2019attributing}, inconsistent iris colors~\cite{li2018detection, mccloskey2018detecting}, {\em etc.} More recently, StyleGAN2~\cite{karras2020analyzing} has greatly improved the visual quality and pixel resolution, with largely-reduced or undetectable artifacts in the generated faces.

{\bf GAN-generated face detection.}
With the development of the GAN models for face generation/synthesis, methods for distinguishing GAN-generated faces have progressed accordingly as well. Most of these methods are Deep Learning based~\cite{marra2019incremental,hulzebosch2020detecting,wang2020cnn,goebel2020detection,liu2020global}. 
Notably, several methods exploit the physiological cues (which suggest inconsistency in the physical world) to distinguish GAN-generated faces from the real ones~\cite{matern2019exploiting}. 
In \cite{yang2019exposinggan}, GAN-generated faces are identified by analyzing the distributions of the facial landmarks. 
The work of \cite{hu2021exposing} analyzes the light source directions from the perspective distortion of the locations of the specular highlights of the two eyes. 
Such physiological/physical-based methods come with intuitive interpretations and are more robust to adversarial attacks~\cite{verdoliva2020media, hu2021tkml}. More related works can be found in the recent survey paper \cite{wang2022gan}.




{\bf Iris and pupil segmentation} is an important task in biometric identification that has been studied well. The IrisParseNet~\cite{wang2020iris} provides complete iris segmentation solutions including iris mask and inner and outer iris boundaries extraction, which are jointly modeled in a unified multi-task neural network. Iris segmentation in non-cooperative environments is supported, while the iris pixel quality might be low due to the limited user cooperation (moving camera, poor illumination, or long-distance views). An end-to-end trainable lightweight stacked hourglass network is presented in \cite{wang2020lightweight} for iris segmentation from noisy images acquired by mobile devices. More recent methods can be found in the NIR Iris Challenge survey paper~\cite{cywangIJCB}.
 
\begin{figure}[t]
\centering
\includegraphics[width=.48\textwidth]{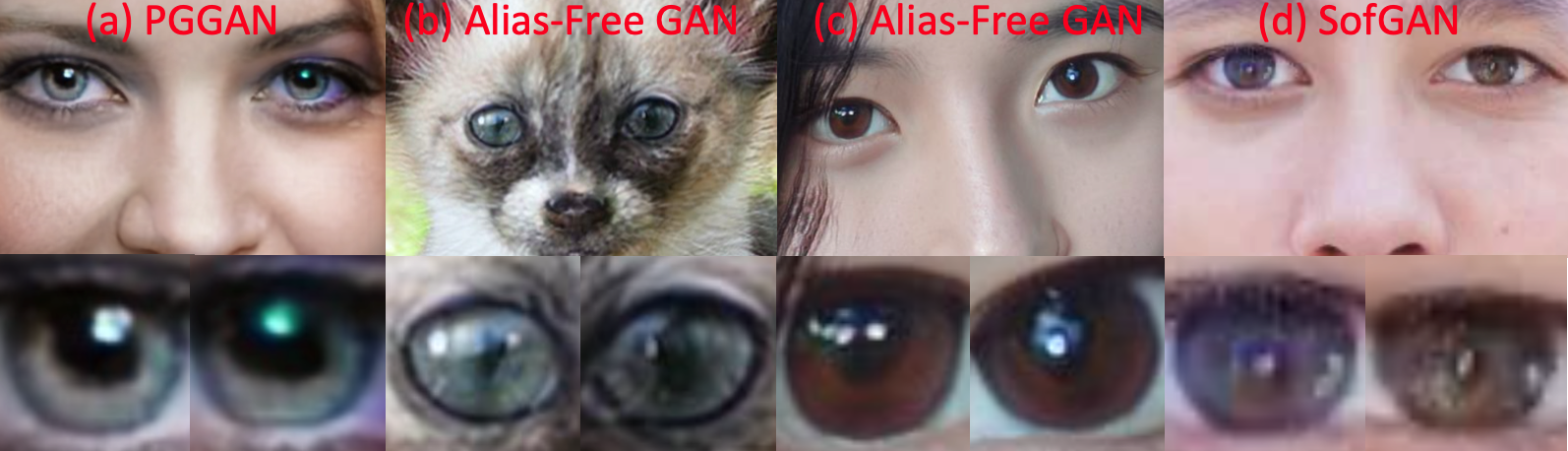}
\vspace{-2em}
\caption{\em Examples of GAN-synthesized faces additional to StyleGAN and StyleGAN2. The images are from their original papers \textbf{(a)} PGGAN \cite{karras2017progressive}, \textbf{(b,c)} Alias-Free GAN (StyleGAN3) \cite{karras2021alias}, \textbf{(d)} SofGAN \cite{chen2021sofgan}. Observe in the zoomed-in view that the pupils appear in irregular, inconsistent shapes, which tell them apart from real faces. }
\vspace{-0.4cm}
\label{fig:other:GANs}
\end{figure}  

\section{Method}
\vspace{-0.2cm}



Our fake face detection method is motivated by the observation that GAN-generated faces exhibit a common artifact that the pupils appear with irregular shapes or boundaries, other than a smooth circle or ellipse. This artifact is universal for all known GAN models (at least for now, {\em e.g.} PGGAN~\cite{karras2017progressive}, Alias-Free GAN~\cite{karras2021alias}, and SofGAN~\cite{chen2021sofgan}), as shown in Figure \ref{fig:other:GANs}. This artifact occurs in both the synthesized human and animal eyes.


Our automatic fake face detection pipeline starts with a face detector to identify any face in the input image. 
We then extract the facial landmark points to localize the eyes and then perform pupil segmentation. The segmented pupil boundary curves are next analyzed to determine if the pupil shape is irregular. We perform parametric fitting of the pupil to an ellipse following the mean squared error (MSE) optimization. This provides a way to define a distance metric to quantify the irregularity for decision-making. The following subsections describe each step of our pipeline in detail.


\subsection{Pupil Segmentation and Boundary Detection} 

We adopt the Dlib~\cite{king2009dlib} face detection to locate the face and extract the 68 facial landmark points provided in Dlib, as shown in Figure \ref{fig:pipeline}. 
We next focus on the eye regions to perform pupil segmentation. 
We use EyeCool~\cite{cywangIJCB}~\footnote{
Code at \url{https://github.com/neu-eyecool/NIR-ISL2021}.
} to extract the pupil segmentation masks with corresponding boundary contours.
EyeCool provides an improved U-Net-based model with EfficientNet-B5~\cite{tan2019efficientnet} as the encoder.
A boundary attention block is added in the decoder to improve the ability of the model to focus on the object boundaries. 
Specifically, considering the subpixel accuracy, we focus on the outer boundary of the pupil 
for the irregularity analysis.






\subsection{Ellipse Fitting to the Pupil Boundary}

\begin{figure}[t]
\centerline{
  \includegraphics[width=0.47\textwidth]{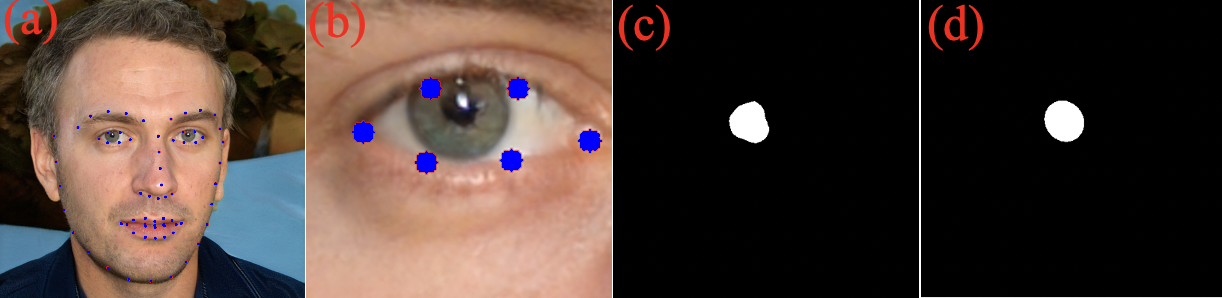}
  \vspace{-0.1cm}
}
\caption{\em The proposed pipeline for face detection, facial landmark localization, pupil segmentation, and pupil ellipse fitting.
\textbf{(a)} The input high-resolution face image,  
\textbf{(b)} The cropped eye image using landmarks, 
\textbf{(c)} The predicted pupil mask of the eye from (b), 
\textbf{(d)} The fitted ellipse mask. 
This example shows a GAN-generated face.
}
\vspace{-1em}
\label{fig:pipeline}
\end{figure}

We next fit an ellipse to the pupil mask via least-square fitting. 
As there might be multiple components in the predicted masks, we keep the largest component for ellipse fitting.
Specifically, the method of \cite{fitzgibbon1999direct} is used to fit an ellipse to the outer boundary of the extracted pupil mask. Figure~\ref{fig:pipeline}(d) shows an example. 
Denotes $u$ as the coordinates of the outer boundary points from the pupil mask. The least-square fitting determines the ellipse parameters $\theta$ minimizing the distance between the pupil boundary points and a parametric ellipse represented by: 
\begin{equation*}
  F(\textbf{u};\mathbf{\theta}) = \mathbf{\theta} \cdot \textbf{u} = ax^2 + bxy + cy^2 + dx +ey +f  = 0,
\end{equation*}
where $\theta = [a,b,c,d,e,f]^T$ and $\textbf{u} = [x^2,xy,y^2,x,y,1]^T$; $T$ denotes transpose. 
$F(\textbf{u};\theta)$ represents the algebraic distance of a 2D point $(x, y)$ to the ellipse, and a perfect fit is indicated by $F(\textbf{u};\theta) = 0$. 
The fitting solution is obtained by minimizing the sum of squared distances (SSD) over the $N$ data points from the pupil boundary:
%
$$
\begin{aligned}
\min_\theta\mathcal{D (\theta)} := \sum_{i=1}^{N} F(u_i;\theta_i)^2, ~~
\text{ s.t.} ~~  ||\theta||^2 =1,  ~~ b^2\ge ac, 
\end{aligned}
$$
%
where the constraints are imposed to avoid the trivial solution of $\theta = \textbf{0}$ and ensure the {\em positive definiteness} of the quadratic form. The solution is calculated using the gradient-based optimization described in \cite{fitzgibbon1999direct}.

\subsection{Estimating the Pupil Shape Irregularity}


To accurately estimate the irregularity of the segmented pupil boundary and the fitted ellipse, we adopt the Boundary IoU (BIoU)~\cite{cheng2021boundary} as a distance metric. 
BIoU is widely used in image segmentation where the sensitivity of the object boundary is important. 
Instead of considering all pixels, BIoU calculates the IoU for mask pixels within a certain distance from the boundary contours between the predicted mask and the corresponding ground truth mask. Thus, BIoU can better focus on the matching of the boundaries of the two shapes. We use BIoU to evaluate the pupil mask pixels that are within a distance of $d$ pixels from the pupil boundary. For each extracted pupil mask, we use $P$ to indicate the predicted pupil mask and $F$ for the fitted ellipse mask. Denote $P_d$ and $F_d$ the mask pixels within distance $d$ from the predicted and fitted boundaries, respectively. BIoU is calculated as: 
\begin{equation}
  \text{BIoU}~  (F, P)   = \frac{|(F_d \cap F)\cap(P_d \cap P )|}{|(F_d \cap F)\cup(P_d \cap P )|}.
 \label{eqbiou}
\end{equation}
The distance parameter $d$ controls the sensitivity of the BIoU measure to the object boundary. Reducing the value of $d$ causes the fitting to be more sensitive to the boundary pixels while ignoring the interior pixels of the pupil mask. 
We set $d=4$ for the BIoU calculation, which leads to the best empirical segmentation performance in our experiments. 

Given the predicted pupil mask and the ellipse fitted pupil mask, 
the BIoU score takes range in $[0,1]$. A larger BIoU value suggests the pupil boundary better fits the parametrized ellipse. In our case, higher BIoU values suggest more regular pupil shapes, and thus the face is more likely real. In comparison, GAN-generated faces should produce lower pupil BIoU scores.    






\begin{figure*}[t]
    \centering
  \includegraphics[width=.49\textwidth]{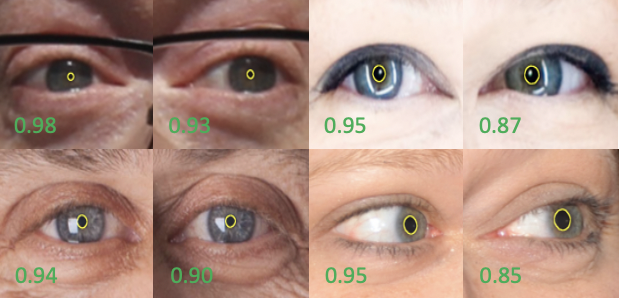}
  \rulesep
  \includegraphics[width=.49\textwidth]{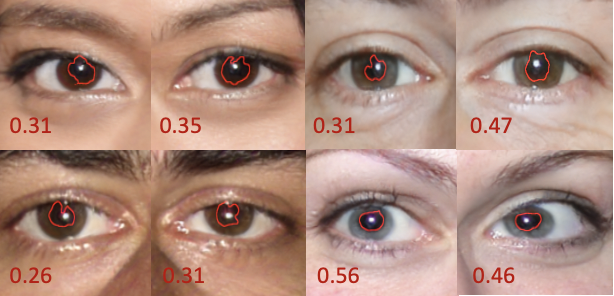}
    \vspace{-1em}
    \caption{\em Examples of both eyes from the real human faces (\textbf{left}) and GAN generated human faces (\textbf{right}). The pixels of the predicted pupil mask within distance $d=4$  from the prediction boundary contours are highlighted. The BIoU scores with $d=4$ between the predicted pupil mask and the ellipse-fitted one are shown on each image. 
    }
    \label{fig:eye_examples}
     \vspace{-1em}
\end{figure*}

\begin{figure*}[t]
\begin{subfigure}[t]{0.30\linewidth}
    \includegraphics[width=\linewidth]{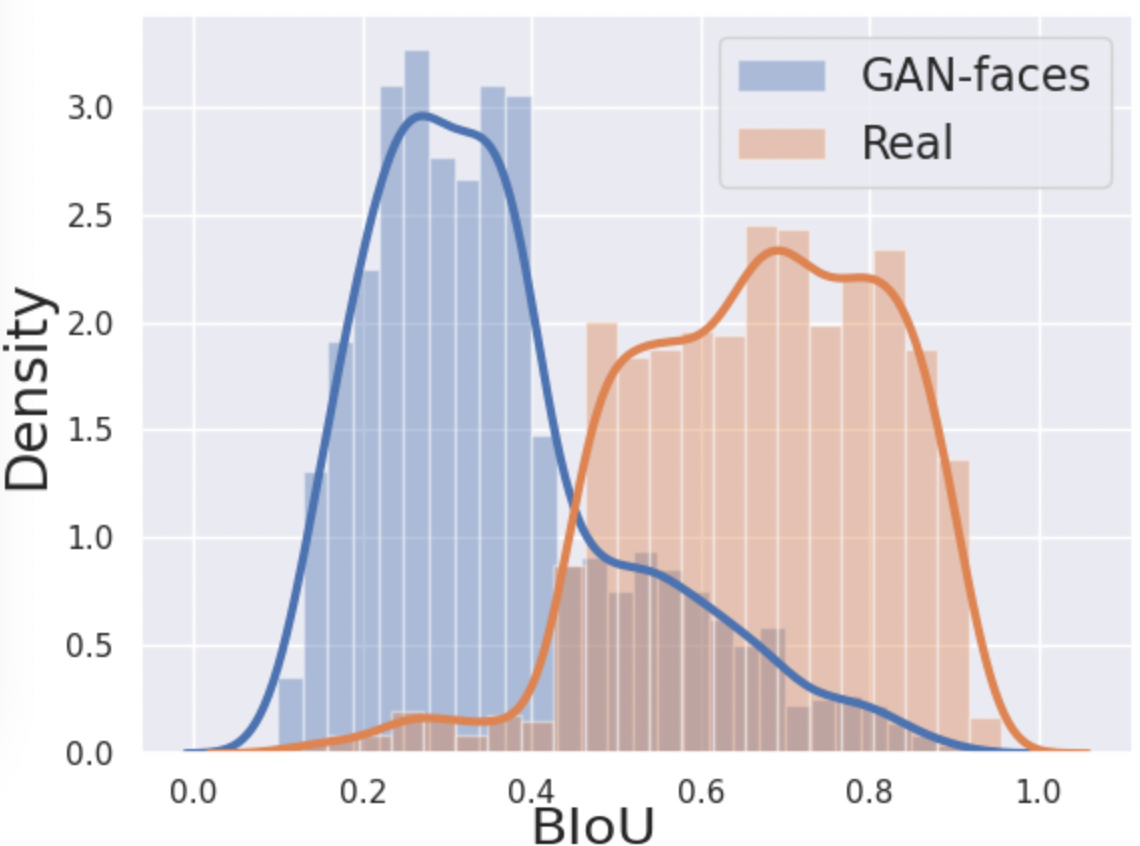}
\end{subfigure}%
    \hfill%
\begin{subfigure}[t]{0.3\linewidth}
    \includegraphics[width=\linewidth]{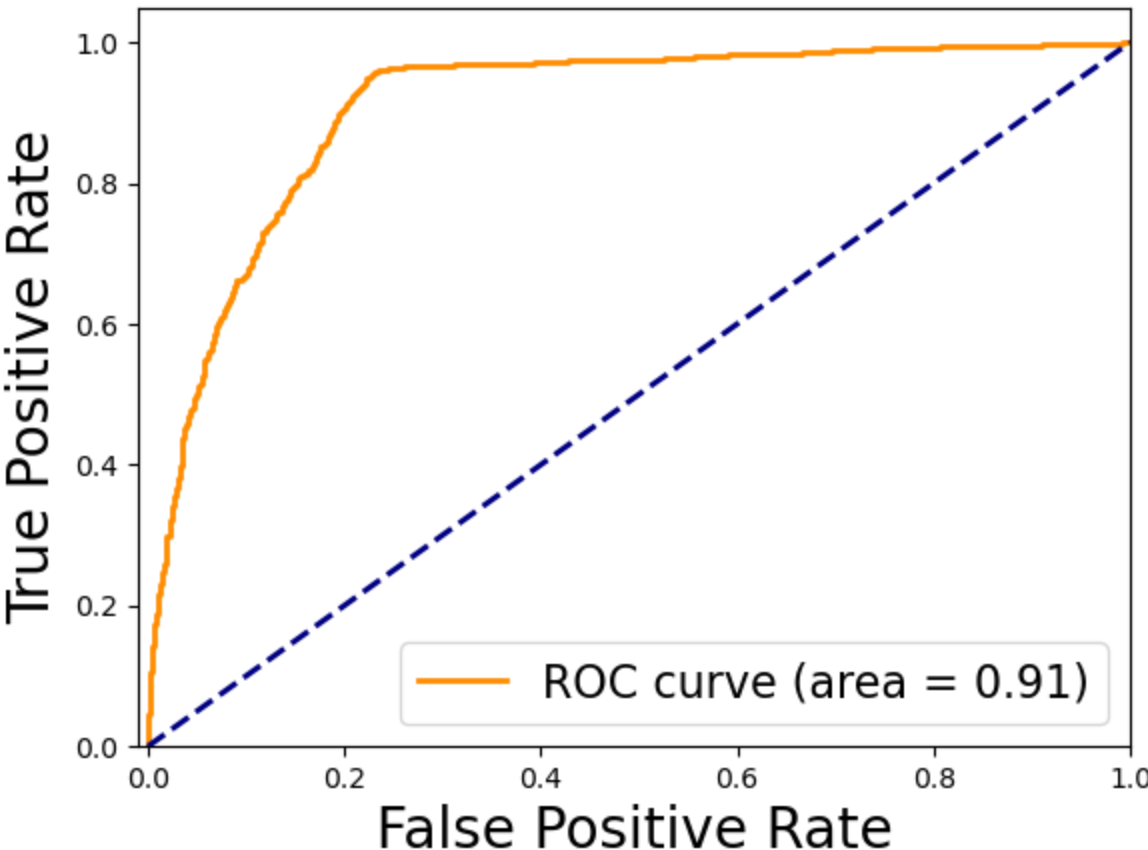}
\end{subfigure}
    \hfill%
\begin{subfigure}[t]{0.3\linewidth}
    \includegraphics[width=\linewidth]{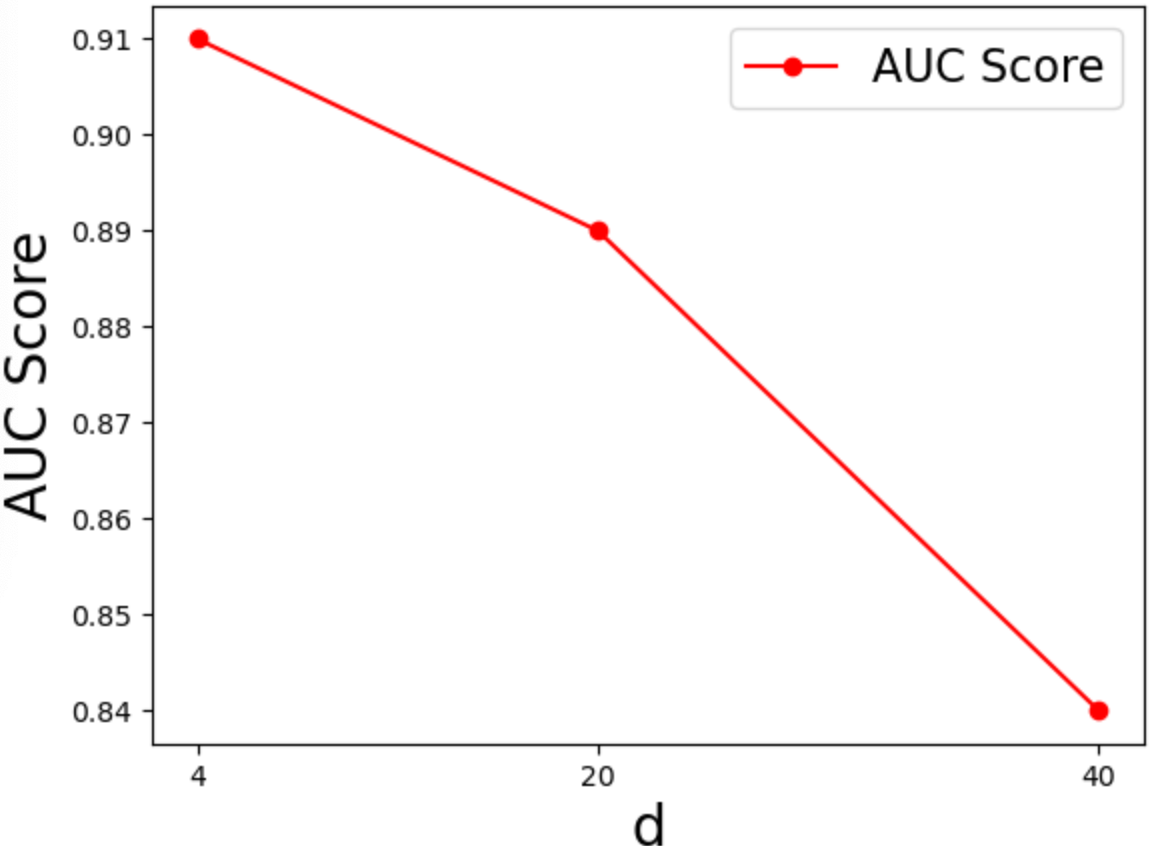}
\end{subfigure}
  \vspace{-0.5em}
\caption{\em  \textbf{Left:} Distributions of the BIoU scores (of the averages of both pupils) for the real and GAN-generated faces. \textbf{Middle:} The ROC curve based on the BIoU with $d=4$. \textbf{Right:} BIoU hyper-parameter analysis, where the $x$-axis indicates distance parameter $d$, and $y$-axis indicates the AUC score.}
\vspace{-1.5em}
\label{fig-eva}
\end{figure*}

\section{Experiments}
\label{sec:experiments}
\vspace{-0.2cm}


\textbf{Datasets.} 
We use the real human faces from the Flickr-Faces-HQ (FFHQ) dataset~\cite{karras2019style}. 
Since StyleGAN2~\cite{karras2020analyzing}~\footnote{\url{http://thispersondoesnotexist.com}} is currently the state-of-the-art GAN face generation model with the best synthesis quality, we collect GAN-generated faces from it. We only use images where the eyes and pupils can be successfully extracted. In total, we collected $1,600$ images for each class (of real {\em vs.} fake faces) with a resolution of $1,024 \times 1,024$.  




\textbf{Results.} 
Figure~\ref{fig:eye_examples} shows examples of the segmented pupils for both the real and GAN-generated faces. These results clearly show that pupils in the real faces are in strongly regular, elliptical shapes.
Such high pupil shape regularity is also reflected in the high BIoU scores computed for the pupil mask and the fitted ellipse. On the other hand, irregular pupil shapes lead to significantly lower BIoU scores, which represents the artifacts of GAN-generated faces. 

Figure~\ref{fig-eva}(left) shows the distributions of the BIoU scores of pupils from the real faces and GAN-generated ones. 
Observe that there is a clear separation between the two classes of distributions, indicating that the proposed {\em pupil shape regularity} can indeed serve as an effective feature to distinguish the GAN-generated faces from the real ones. 

Figure~\ref{fig-eva}(middle) shows the {\em receiver operating characteristic} (ROC) curve of our GAN-generated face detection evaluation. The Area under the ROC curve (AUC) score is 0.91, which indicates the effectiveness of the proposed method. 

\textbf{Sensitivity Analysis of $d$.}
The BIoU boundary distance $d$ is an essential parameter that controls the matching sensitivity of the pixels near shape boundary. Figure \ref{fig-eva}(right) shows how the fake face detection ROC varies {\em w.r.t.} parameter $d$. 
As the value of $d$ grows too large, sensitivity telling the differences of pupil boundary decreases, which also reduces fake face detection performance.


\begin{figure}[t]
\centering
\includegraphics[width=.23\textwidth]{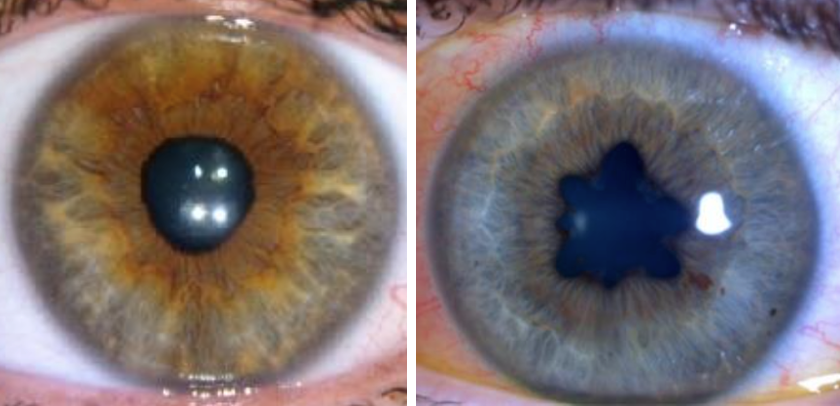}
\rulesep
\includegraphics[width=.225\textwidth]{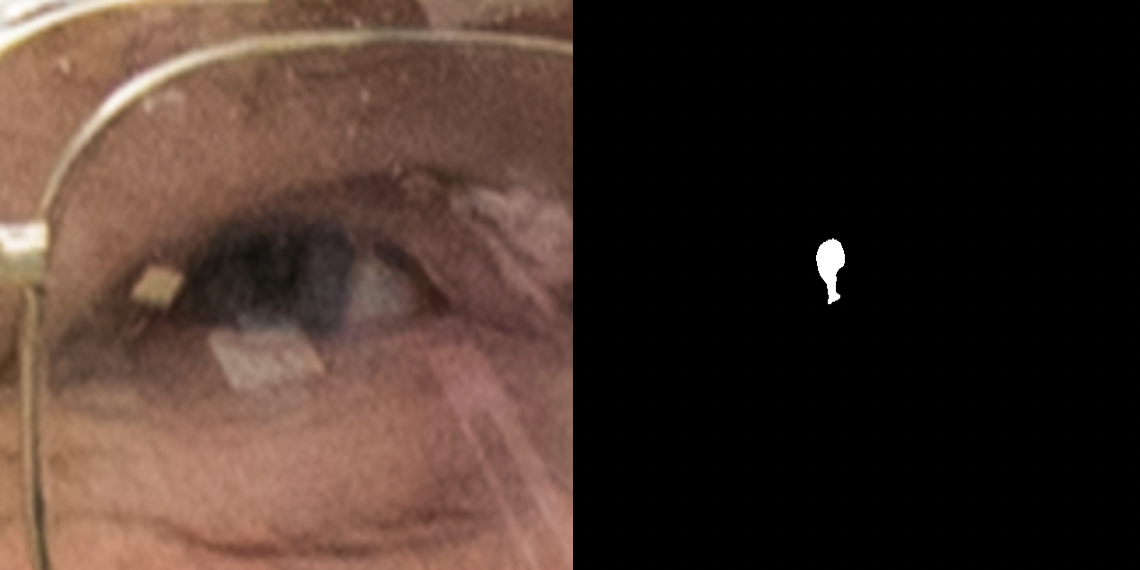}
\vspace{-1em}
\caption{\em 
\textbf{Left:} 
Examples of diseased and infection eyes from \cite{AbnormalEye2017}.    
These pupils from images of real faces contain abnormal non-elliptical pupil shapes, which only occurs rarely in real life. 
\textbf{Right:} Occlusions and environmental variations can cause pupil segmentation failure.}
\vspace{-2em}
\label{fig:abeye} 
\end{figure}  

\textbf{Limitations.} 
The proposed method still contains several limitations. Since our method is based on the simple assumption of pupil shape regularity, false positives may occur when the pupil shapes are non-elliptical in the real faces. This may happen for infected eyes of certain diseases as shown in Figure \ref{fig:abeye}(left). Also imperfect imaging conditions including lighting variations, largely skew views, and occlusions can also cause errors in pupil segmentation or thresholding errors, as shown in Figure \ref{fig:abeye}(right). 



\section{Conclusion}
\vspace{-0.2cm}

In this paper, we show that GAN-generated faces can be identified by exploiting the regularity of the pupil shapes. We propose an automatic method for pupil localization and segmentation, and perform ellipse fitting to the segmented pupils to estimate a Boundary IoU score for forensic classification. The proposed approach is simple yet effective. The detection results are interpretable based on the BIoU score.

{\bf Future Work.} We will investigate other types of inconsistencies between two pupils of the GAN-generated face, such as the different geometric shapes and relative locations of pupils in the two eyes. These cues in combination may further improve forensic detection effectiveness. Future work also includes the deployment to an online platform that can further expand the impact in addressing issues in social media forensics.

\begin{figure*}[h!]
     \centering
     \includegraphics[width=.95\textwidth]{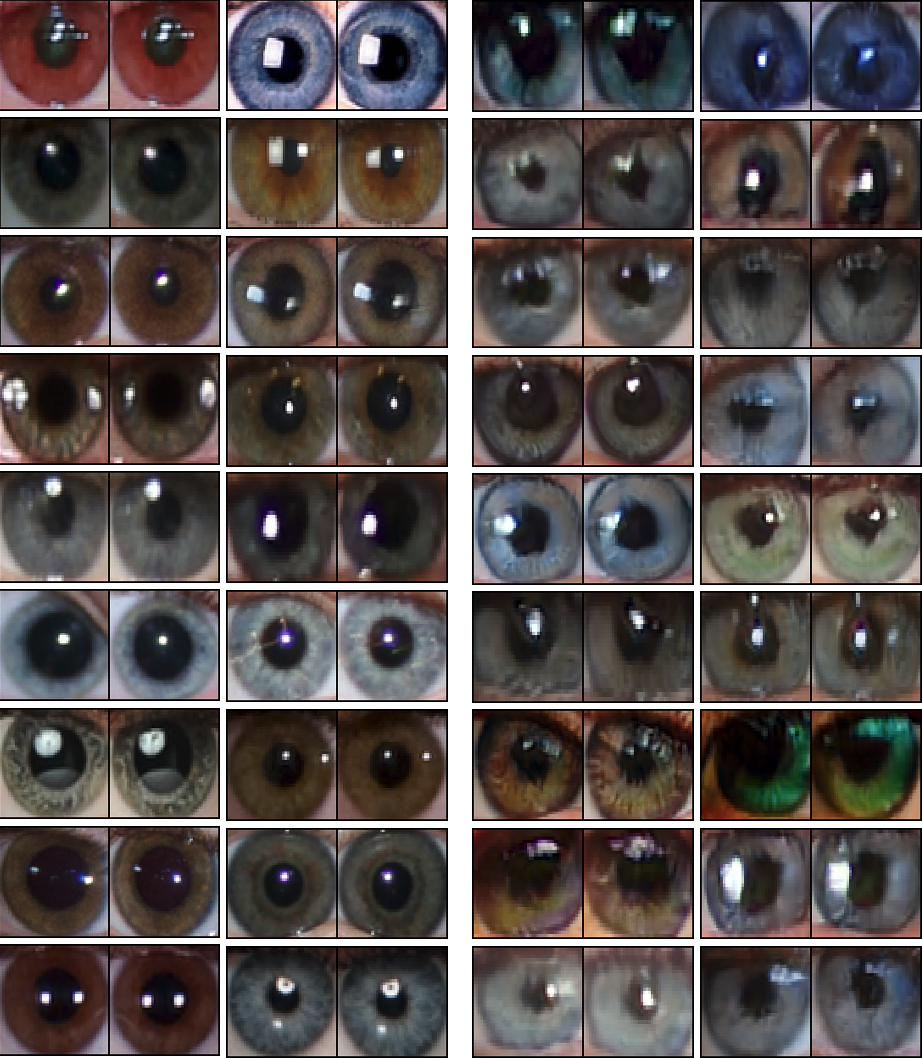} 
     \caption{\em More pair of pupil examples from real (\textbf{left}) human faces and GAN-generated (\textbf{right}) faces. As we mentioned before, the irregular pupil shape is a good sign for the human to expose the GAN-generated face visually, even there are no boundary labels around the pupils, we can easily see that the shapes of GAN-generated pupils are very irregular, and the shapes of both pupils are very different in the same GAN-generated face image. In practice, people can zoom a face image large enough and then check the pupil shapes to find whether the face is real or not easily. }
     \label{fig-humancheck}
 \end{figure*}



{\small
\bibliographystyle{IEEEbib}
{\raggedright
\bibliography{main}
}
}


\end{document}